\documentclass[runningheads]{llncs}
\usepackage[T1]{fontenc}
\usepackage{graphicx}
\usepackage[style=lncs,backend=biber]{biblatex}
\usepackage{booktabs}
\usepackage{amsmath, amssymb}
\begin{document}
\title{F\textsuperscript{2}IND-IT! -  Multimodal Fuzzy Fake Indian News Detection using Images and Text}

\titlerunning{F\textsuperscript{2}IND-IT! - Multimodal Indian Fuzzy Fake News Detection}

\author{
Kushal Trivedi\inst{1} \and
Murtuza Shaikh\inst{1} \and
Khushi Singh\inst{1} \and
Jeevaraj S.\inst{1}
}

\authorrunning{Kushal et al.}

\institute{
ABV - Indian Institute of Information Technology, Gwalior \\
\email{kushal.trivedi.2110@gmail.com}
}

\maketitle              
\begin{abstract}
Newspapers remain a vital source of journalism, delivering updates on current events, politics, business, sports, and entertainment. However, in a country as vast and diverse as India, partial or biased manipulation  of facts is common, especially when the same news is covered by multiple regional and national outlets. While several existing approaches focus on integrating textual and visual features for fake news detection, very few have examined their effectiveness on Indian news content. This research presents a novel multimodal framework — F\inst{2}IND-IT! (Fuzzy Fake Indian News Detection using Images and Text) — that combines visual and textual modalities for enhanced fake news detection on Indian media. The proposed model utilizes Convolutional Neural Networks (ResNet-50) to extract visual features from news images, a text encoder (DistilBERT) to obtain textual semantic embeddings and an Adaptive Neuro-Fuzzy Inference System (ANFIS) to generate a fuzzy reliability score. A lightweight attention-based fusion module is employed to assign learnable weights to each modality before classification into fake or real. The study is completed by a formal and in-depth analysis and exploration of this new architecture to the IFND dataset with comparison to previous research by comparing accuracy, precision, recall and F1 scores, to support a brief discussion on the model’s performance.

\keywords{Fuzzy \and Multimodality\and Fake News.}
\end{abstract}

\section{Introduction}
With the advancement, awareness, and wider use of technology in recent years, the number of news articles being shared has increased sharply. Earlier, daily news was mostly available through newspapers and reached only a small part of the population. According to a report by the Indian Ministry of Communication in March 2024, 95.15\% of villages in India now have 3G or 4G mobile internet access \cite{PIB2025Varanasi}. In addition, reports by IAMAI (2024) predict that by 2025, about 56\% of all new internet users in India will come from rural areas \cite{CommToday2024RuralInternet}.

However, as news has become more accessible to more people, the amount of fake news has also grown. While there is no official definition of "fake news," it is commonly described as \textit{any content that is deliberately made and known to be false}. Fake news often uses emotional language and special writing styles. These are often captured through features such as tone and writing patterns. Common ways to spread fake news include editing images, changing topics to mislead readers, and using clickbait to attract attention \cite{Tufchi2023MultimodalFakeNews}.

According to official data from the Press Information Bureau under the Ministry of Information and Broadcasting, 1,575 fake news cases were reported between 2022 and March 2025. The number rose from 338 in 2022 to 583 in 2024 \cite{PIB2025_FakeNewsCases}. Data from the National Crime Records Bureau also shows a 214\% increase in fake news cases during the early pandemic period from 2018 to 2020 \cite{NCRB2021_FakeNewsRise}. A 2024 study by ISB and CyberPeace found that 46\% of false information was about politics, and over 77\% of it spread through social media platforms \cite{NDTV2024_ISBFakeNews}. Another survey among Gen Z users in Delhi found that 91\% believe fake news can affect election outcomes \cite{brandequity2024fakenews}.

Current methods for detecting fake news automatically are usually grouped into three types: modality-based methods, propagation-based methods, and fact-based methods \cite{10889602}. Modality-based methods look at the content of the news itself. This includes text features such as writing style or word use, image features such as signs of editing, or both text and image combined (multimodal). Propagation-based methods study how news spreads across social media and other online platforms. Fact-based methods try to check the news content against trusted sources or known facts.

The evolution from single-modal to multi-modal fake news classifiers has become essential to prevent underfitting models trained on only one type of content modality, such as either textual or visual data. In many cases, the information conveyed through images may contradict the textual content, or vice versa, which can lead to misleading interpretations and ultimately contribute to biased news diffusion. Multi-modal approaches aim to capture complementary features from both text and image modalities, enabling more robust and accurate detection of fake news.

The World Economic Forum’s 2024 Global Risk Report placed India at highest risk for misinformation globally, with experts citing high levels of political polarization and algorithmic amplification \cite{WEF2024Risks}.  Manual detection of fake news is labor-intensive, time-consuming, and prone to bias too. Thus, there is a significant gap in the creation of a credible database centered around Indian news articles, as well as in research focused on developing tools for robust automated classification of fake news. This gap serves as the motivation for introducing F\inst{2}IND-IT, a fuzzy-based multimodal deep learning architecture for fake news detection.

\section{Prior Art}

\subsection{Baseline Deep Learning Approaches for Multi-Modal Data}
MAGIC \cite{xumultimodal}, IFND \cite{Sharma2023}, Tri‑FusionNet\cite{10770746}, BDANN \cite{9206973}, CLIP‑based learning \cite{10219997}, Cross‑Attention Networks \cite{9541113}, and ETMA \cite{10077443} are among the top-performing frameworks in multimodal fake‑news detection. Table~\ref{tab:dl_multimodal} summarizes their accuracy, F1 scores, and datasets used.

\begin{table}[h]
\centering
\caption{Summary of top-performing deep learning frameworks for multimodal fake news detection}
\label{tab:dl_multimodal}
\small
\begin{tabular*}{\textwidth}{@{\extracolsep{\fill}}llcc@{}}
\toprule
\textbf{Model} & \textbf{Dataset(s)} & \textbf{Acc. (\%)} & \textbf{F$1$-Score} \\
\midrule
MAGIC & Fakeddit, Chinese MM & 98.72 / 85.96 & 0.98 / 0.97 \\
IFND & IFND (Indian) & 74.00 & --- \\
Tri-FusionDet & Weibo, Politifact & 94.00 & 0.94 \\
BDANN & Weibo & 86.90 & 0.85 \\
CLIP-based & Weibo, Politifact, Gossipcop & 90.70 / 94.20 / 88.00 & 0.90 / 0.89 / 0.63 \\
Cross-attention & Weibo, Pheme & 87.90 / 87.20 & 0.88 / 0.85 \\
ETMA & Twitter, Pontes, Jruvika, Risdal & 93 / 96 / 97 / 95 & 0.92 / 0.94 / 0.95 / 0.96 \\
\bottomrule
\end{tabular*}
\end{table}

\subsection{Fuzzy-based Deep Learning Approaches for Multi-Modal Data}

To the best of our knowledge, \cite{GEDARA2025113277} is the only work that incorporates fuzzy logic with neural networks for fake news classification. The results from this study are summarized in Table~\ref{tab:dl_fuzzy}.

\begin{table}[h]
\centering
\caption{Summary of performance of the neuro-fuzzy model.}
\label{tab:dl_fuzzy}
\begin{tabular*}{\textwidth}{@{\extracolsep{\fill}}lcc@{}}
\toprule
\textbf{Dataset} & \textbf{Accuracy (\%)} & \textbf{F1 Score} \\
\midrule
Twitter    & 94.10 & 0.921 \\
BuzzFeed   & 89.30 & 0.902 \\
PolitiFact & 90.30 & 0.901 \\
\bottomrule
\end{tabular*}
\end{table}



\section{Proposed Methodology}

In this section, we present the methodology of the proposed framework, including the dataset used, various CNN architectures, text encoders, and the experimental setup.

\subsection{Dataset Used}
In this study, we consider IFND (Indian Fake News Dataset). The IFND (Indian Fake News Dataset) is a multimodal dataset containing image-text pairs extracted from Indian news articles. It comprises 56,713 news articles, covering international, national, and local events from the period between 2013 and 2021. This dataset has been used in our study to classify fake news from real news. The articles are categorized into five topics—Election, Politics, COVID-19, Violence, and Miscellaneous.

\subsection{F\inst{2}IND-IT Architecture}
\begin{figure}[p]
\centering
\includegraphics[scale=0.5]{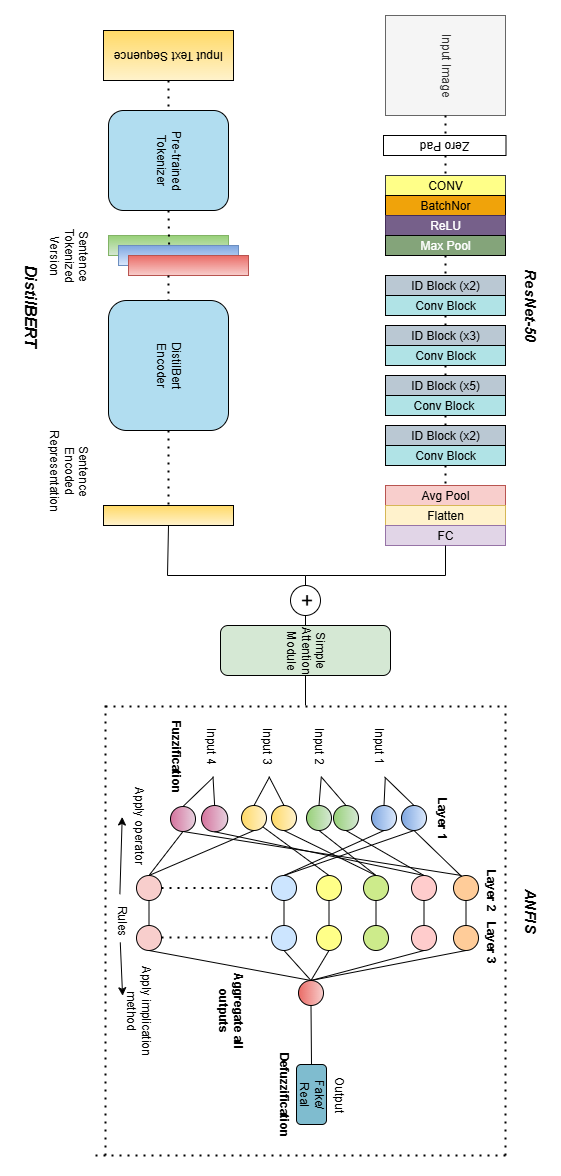}
\caption{The framework of the F\textsuperscript{2}IND Architecture.}
\label{fig3}
\end{figure}

In this subsection, we discuss the overall architecture of the proposed F\inst{2}IND-IT model (illustrated in Fig.~\ref{fig3}). 
\subsubsection{Overall Flow of Data}

This model uses DistilBERT and ResNet-50 to extract textual and visual features, respectively, projecting them into high-dimensional embeddings. A lightweight attention gating mechanism fuses the modalities, with embeddings resized via MLPs for dimensional alignment. The attention module adaptively balances each modality’s contribution. The fused features are then passed through an ANFIS layer with 2 Gaussian membership functions to perform binary fake-news classification.

\subsubsection{Visual Feature Extractor (CNN)}

We utilize a ResNet-50-based visual encoder to extract high-level image features. Specifically, we load the pretrained ResNet-50 model and remove its final classification layer, retaining only the convolutional backbone. Formally, given an input image $I$, the encoder maps it to a fixed-size feature vector:
\[
v = \text{ResNet}(I) \in \mathbb{R}^{2048},
\]
where $v$ represents the output of the global average pooling layer. All parameters of the ResNet backbone are fine-tuned during training to better align with the target task.

\subsubsection{Text Encoder (DistilBERT)}
For a piece of news, we use the \texttt{DistilBert-\\Tokenizer} to tokenize its contents, adding the classification token \texttt{[CLS]} at the beginning and the separation token \texttt{[SEP]} at the end of the token sequence. The resulting input takes the form:
\[
X = [\texttt{[CLS]}, x_1, \ldots, x_n, \texttt{[SEP]}],
\]
where $n$ is the number of original tokens. These tokens are then fed into DistilBERT, which maps them into a contextualized low-dimensional embedding space:
\[
W = \text{DistilBERT}(X) \in \mathbb{R}^{N \times d},
\]
where $d = 768$ is the hidden size of the \texttt{distilbert-base-uncased} model. Unlike the original BERT, DistilBERT removes the token-type embeddings and the second segment input, offering a more lightweight and faster alternative while retaining 95\% of BERT’s language understanding capabilities.

In our implementation, the output of the DistilBERT encoder is a tensor of shape $(B, S, 768)$, where $B$ is the batch size and $S$ is the sequence length. To obtain a fixed-size sentence representation, we apply a mean pooling operation over the token embeddings, weighted by the attention mask. Specifically,
\[
\mathbf{w}_{\text{mean}} = \frac{\sum_{i=1}^{N} \mathbf{w}_i \cdot m_i}{\sum_{i=1}^{N} m_i},
\]

where $\mathbf{w}_i$ is the hidden representation of the $i$-th token and $m_i$ is the corresponding attention mask. This results in a single vector per input sequence of shape $(B, 768)$, which serves as the final sentence embedding.
\subsubsection{Attention-Based Fusion Module}
The shapes of tensors from the ResNet-50 module and DistilBERT encoder are established as X=2048 and Y=768, respectively. Before feeding embeddings into the ANFIS module for the fuzzy inference implementation of the model, both embeddings are projected to a common dimensional space of size 512. After projection, we stack them along the modality dimension, resulting in a combined tensor of shape $(B, 2, 512)$, where $B$ denotes the batch size.

To compute attention logits for each modality, we apply an MLP that projects each modality-specific embedding to a scalar value, producing attention scores of shape $(B, 2)$. These scores are then renormalized (via softmax) to ensure they sum to 1 across modalities. The embeddings are then aggregated and reshaped back to a unified representation of shape $(B, 512)$, which is subsequently used for the final binary classification task. These steps can be represented mathematically as:

\begin{equation}
\begin{aligned}
    &x \in \mathbb{R}^{B \times 2048}, \quad y \in \mathbb{R}^{B \times 768} \\
    &\hat{x} = W_x x \in \mathbb{R}^{B \times 512}, \quad \hat{y} = W_y y \in \mathbb{R}^{B \times 512} \\
    &z = [\hat{x}; \hat{y}] \in \mathbb{R}^{B \times 2 \times 512} \\
    &a = \text{softmax}(\text{MLP}(z)) \in \mathbb{R}^{B \times 2} \\
    &h = \sum_{i=1}^{2} a_i \cdot z_i \in \mathbb{R}^{B \times 512}
\end{aligned}
\label{eq:fusion}
\end{equation}

\subsubsection{Fuzzy Logic Inference (ANFIS)}
\begin{enumerate}
    \item \textbf{Input Layer:} The input to ANFIS consists of batches of 4-dimensional vectors, i.e., of shape $(B, n)$ where $n = 4$. Let $X = \{x_1, x_2, x_3, x_4\}$ represent the input features.\\

    \item \textbf{Fuzzification Layer:} Each input value is fuzzified using two Gaussian membership functions. The mean ($\mu_j$) and standard deviation ($\sigma_j$) of each membership function are learnable parameters.\\

    For every feature $x_i$ in the input $X$, the degree of membership to each fuzzy set is computed using the Gaussian function:
    \[
    G(x_i; \mu_j, \sigma_j) = \exp\left(-\frac{(x_i - \mu_j)^2}{2\sigma_j^2}\right),
    \]
    where $i \in [1, n]$ and $j \in [1, f]$, with $n = 4$ and $f = 2$ (number of membership functions). Thus, for each input $X$, $n \times f = 4 \times 2 = 8$ membership values are computed. The output of this layer is of shape $(B, n, f)$. \\

    \item \textbf{Rule Layer:} All possible fuzzy rules are evaluated using the product (AND) of the membership values across features. The total number of fuzzy rules is $f^n = 2^4 = 16$.\\

    The firing strength $f_k$ of the $k$-th rule is computed as:
    \[
    f_k = \prod_{i=1}^{n} G(x_i; \mu_j, \sigma_j),
    \]
    where $k \in [1, f^n]$.\\

    The firing strengths are then normalized:
    \[
    \hat{f}_k = \frac{f_k}{\sum_{i=1}^{f^n} f_i}.
    \]
    The output of this layer is of shape $(B, f^n)$.

    \item \textbf{Rule Weighting Layer:} Each rule contributes a weighted output, computed as:
    \[
    z_k = \sum_{i=1}^{n} a_{ik} x_i + b_k,
    \]
    where $a_{ik}$ and $b_k$ are trainable parameters for each rule and input feature.

    The output of this layer is also of shape $(B, f^n)$.\\

    \item \textbf{Output Layer:} The final output is a weighted sum of the normalized firing strengths and the rule outputs:
    \[
    z = \sum_{k=1}^{f^n} \hat{f}_k \cdot z_k.
    \]

    A sigmoid activation is applied to produce a confidence score between 0 and 1:
    \[
    \text{Output} = \sigma(z) = \frac{1}{1 + e^{-z}}.
    \]

    The final output is of shape $(B, 1)$, representing the fake news probability for each input in the batch.
\end{enumerate}

\subsection{Evaluation Metric}
F1 score metric is a comprehensive evaluation method of precision and recall, we take it as the metric in evaluating our approach and baselines. Recall, Precision, and F1 equations are shown as follows:

\[
F1 = \frac{2 \cdot \text{Precision} \cdot \text{Recall}}{\text{Precision} + \text{Recall}}
\]

\subsection{Experiment Setting}
The details of the experimental setup of our approach are discussed below:

\begin{enumerate}
    \item \textbf{Data Imbalance and Preprocessing:}  
    The dataset consists of 56,713 news article text-image pairs. However, due to missing image links, a significant number of samples were removed for image preprocessing. All images with a resolution higher than $224 \times 224$ were resized to $224 \times 224$ and normalized using the ImageNet dataset statistics. This resulted in a final dataset of $25,195$ examples, comprising $24,576$ true news articles and $619$ fake news articles. Dynamic padding is also used for text batch preprocessing.

    \item \textbf{DistilBERT:}  
    A dropout rate of 0.30 was applied, and mean pooling was used to obtain fixed-size sentence embeddings.

    \item \textbf{ResNet-50:}  
    The final classification layer was removed, and all model parameters were kept trainable to enable fine-tuning.

    \item \textbf{Attention Fusion:}  
    A modality-level attention mechanism was implemented using a lightweight MLP and softmax normalization to compute attention scores between modalities. Bit-masking is applied to examples where images are unavailable, ensuring that the attention fusion mechanism allocates complete attention to the text, effectively setting the image's weight to 0.

    \item \textbf{ANFIS:}  
    A Takagi–Sugeno–style Adaptive Neuro-Fuzzy Inference System (ANFIS) is employed utilizing $4$ inputs and $2$ Gaussian membership functions.

    \item \textbf{Loss, Learning Rate, and Optimizer:}  
    A custom loss function was designed as a weighted combination of binary cross-entropy loss, Huber loss (to penalize incorrect minority class predictions), and focal loss (to address class imbalance). The learning rate was dynamically adjusted using the OneCycleLR scheduler, with different scales assigned to different model components. The Adam optimizer was used for all updates.
\end{enumerate}

Furthermore, the model was trained using a stratified 5-fold cross-validation strategy for 5 epochs, with a batch size of 16. The number of parameters in the model are roughly 91.3 million.

\section{Results and Discussion}

\begin{table}[h]
\centering
\caption{Performance of the \textit{F$^2$IND} model on the IFND dataset}
\label{tab:results}
\small 
\begin{tabular*}{\textwidth}{@{\extracolsep{\fill}}lcccccccc@{}}
\toprule
\textbf{Model} & \textbf{Dataset} & \textbf{Acc.} & \multicolumn{2}{c}{\textbf{Macro-F1}} & \textbf{Prec.} & \textbf{Recall} & \textbf{ROC} & \textbf{PR} \\
\cmidrule(lr){4-5}
& & & \textit{Fake} & \textit{True} & & & \textbf{AUC} & \textbf{AUC} \\
\midrule
F$^2$IND & IFND & \textbf{0.9769} & 0.9735 & 0.9789 & 0.9801 & 0.9854 & \textbf{0.9959} & \textbf{0.9977} \\
\bottomrule
\end{tabular*}
\end{table}

\begin{enumerate}
    
    \item \textbf{Evaluation Metrics:} To counter class imbalance, Macro F1 is used over Micro F1. This ensures equal weighting across all classes and prevents majority-class bias.
    
    \item \textbf{Validation:} K-Fold Cross-Validation was implemented to ensure robust evaluation and improve generalization by rotating training/validation sets across all data points.
    
    \item \textbf{Fuzzy Membership:} Gaussian membership functions exhibit moderate overlap for smooth transitions and distinct centers, effectively capturing non-linear trends.
    
    \item \textbf{Firing Strength and Rule Contributions:} Most of the 16 rules show uniform normalized firing strengths. This indicates a balanced system architecture. The model’s output is driven by a subset of rules with both positive and negative contributions, allowing for precise pattern discrimination and prediction.
\end{enumerate}

\section{Ablation Studies}

The following ablation studies (Table~\ref{tab:ablation_optimized}) were conducted in addition to our primary research to evaluate performance improvements of the proposed model. Further ablation studies can also be conducted by varying the number and nature of membership functions, as well as the ANFIS architecture itself.
    
\begin{table}[h!]
\centering
\renewcommand{\arraystretch}{1.2} 
\begin{tabular*}{\textwidth}{@{\extracolsep{\fill}}llc@{}}
\hline
\textbf{Category} & \textbf{Model/Configuration} & \textbf{Accuracy (\%)} \\ \hline
CNN (Unimodal) & ResNet-50 & 76.60 \\
& VGG-16 & 65.30 \\ \hline
Text Encoder (Unimodal) & LSTM & 92.60 \\
& Bi-LSTM & 92.70 \\ \hline
CNN (Multimodal) & ResNet-50 & 97.69 \\
& VGG-19 & 97.85 \\ \hline
Text Encoder (Multimodal) & LSTM & 94.17 \\
& Bi-LSTM & 95.67 \\
& DistilBERT & 97.69 \\
& BERT & 97.71 \\ \hline
ANFIS Influence & Without ANFIS & 96.73 \\
& With ANFIS & \textbf{97.69} \\ \hline
\end{tabular*}
\caption{Ablation study comparing architectures and configurations.}
\label{tab:ablation_optimized}
\end{table}

\section{Conclusion and Future Work}
This research presents a novel study on the detection of fake news published by Indian newspapers and proposes a new architecture that combines neural networks and fuzzy logic to classify news as either fake or real. Experimental results on the real-world, comprehensive IFND dataset demonstrate the effectiveness of the proposed model. Ablation studies were conducted to explore alternative model architectures. While some variations performed marginally close to our proposed architecture, ours consistently outperformed them across all evaluation metrics.

Potential future enhancements to this study include replacing ANFIS's reliance on prior expert knowledge for forming input-output fuzzy partitions and designing the fuzzy rule base with data-driven models that automatically identify the centroids and spreads of fuzzy clusters during training. In these models, rules are formed dynamically during training and do not require expert intervention \cite{KasabovSong2002_DENFIS}\cite{Kasabov2001_EFuNN}\cite{IyerQuek2018_PIE_RSPOP}.

\begin{credits}
\subsubsection{\discintname}
The authors declare that they have no known competing financial
interests or personal relationships that could have appeared to influence the work reported in this paper. Pre-processed dataset and code for the model can be made available upon request.
\end{credits}

\printbibliography
\end{document}